%% file: arxiv.tex
\documentclass[10pt,twocolumn,letterpaper]{article}

\usepackage[pagenumbers]{wacv} 

\input{preamble}

\definecolor{wacvblue}{rgb}{0.21,0.49,0.74}
\usepackage[pagebackref,breaklinks,colorlinks,allcolors=wacvblue]{hyperref}

\def\wacvPaperID{887} 
\def\confName{WACV}
\def\confYear{2026}

\title{Q-Former Autoencoder: A Modern Framework for Medical Anomaly Detection 
}

\author{Francesco Dalmonte$^{1,*}$\\
\and
Emirhan Bayar$^{2,}$\thanks{Equal contribution}\\
\and
Emre Akbas$^{2, 3}$\\
\and
Mariana-Iuliana Georgescu$^{3}$\\
\and
$^1$University of Bologna, Italy \and
$^2$Middle East Technical University, Ankara, Türkiye \and
$^3$Helmholtz Munich, Germany 
}

\begin{document}
\maketitle

\input{sec/0_abstract}    
\input{sec/1_intro}
\input{sec/2_related_work}

\input{sec/3_method}

\input{sec/4_results}
\input{sec/5_conclusion}

{
    \small
    \bibliographystyle{ieeenat_fullname}
    \bibliography{arxiv}
}

\newpage

\maketitlesupplementary

\input{sec/supp}

\end{document}

%% file: preamble.tex
\usepackage{tabularray,xcolor, multirow}
\newcommand{\rownumber}[1]{\textcolor{darkblue}{#1}}
\definecolor{darkblue}{RGB}{0, 112, 188}
\usepackage{pifont} 
\usepackage[table]{xcolor}
\newcommand{\xmark}{\ding{55}}%
\newcommand{\cmark}{\ding{51}}
\definecolor{light}{RGB}{179, 224, 255}
\usepackage{wrapfig}
\newcommand{\rc}{\cellcolor{light}}
\usepackage{svg} 
 

%% file: sec/0_abstract.tex
\begin{abstract}
Anomaly detection in medical images is an important yet challenging task due to the diversity of possible anomalies and the practical impossibility of collecting comprehensively annotated data sets. 
In this work, we tackle unsupervised medical anomaly detection proposing a modernized autoencoder-based framework, \textbf{the Q-Former Autoencoder}, that leverages state-of-the-art pretrained vision foundation models, such as DINO, DINOv2 and Masked Autoencoder.
Instead of training encoders from scratch, we directly utilize frozen vision foundation models as feature extractors, enabling rich, multi-stage, high-level representations without domain-specific fine-tuning. 
We propose the usage of the Q-Former architecture as the bottleneck, which enables the control of the length of the reconstruction sequence, while efficiently aggregating multi-scale features. 
Additionally, we incorporate a perceptual loss computed using features from a pretrained Masked Autoencoder, guiding the reconstruction towards semantically meaningful structures. 
Our framework is evaluated on four diverse medical anomaly detection benchmarks, achieving state-of-the-art results on BraTS2021, RESC, and RSNA. 
Our results highlight the potential of vision foundation model encoders, pretrained on natural images, to generalize effectively to medical image analysis tasks without further fine-tuning. We release the code and models at \url{https://github.com/emirhanbayar/QFAE}.
 
\end{abstract}

%% file: sec/1_intro.tex
\section{Introduction}
\label{sec:intro}

Automated anomaly detection in medical imaging is a crucial problem, as it directly impacts diagnostic accuracy, workflow efficiency, and patient outcomes. 
However, manual inspection of large-volume medical scans, such as Magnetic Resonance Imaging (MRI) or Computed Tomography (CT), is inherently time-consuming and susceptible to human error, highlighting the need for reliable automated systems that can assist physicians in flagging potential anomalies. 
However, automated medical anomaly detection presents significant challenges too. Anomalies manifest in highly diverse forms and appearances, making it infeasible to collect representative samples of all possible pathological variations. 
As a result, unsupervised anomaly detection approaches, which train models exclusively on normal data to identify deviations as anomalies, are appropriate for this domain.  

\begin{figure}[t]
  \centering
  \includegraphics[width=0.99\linewidth]{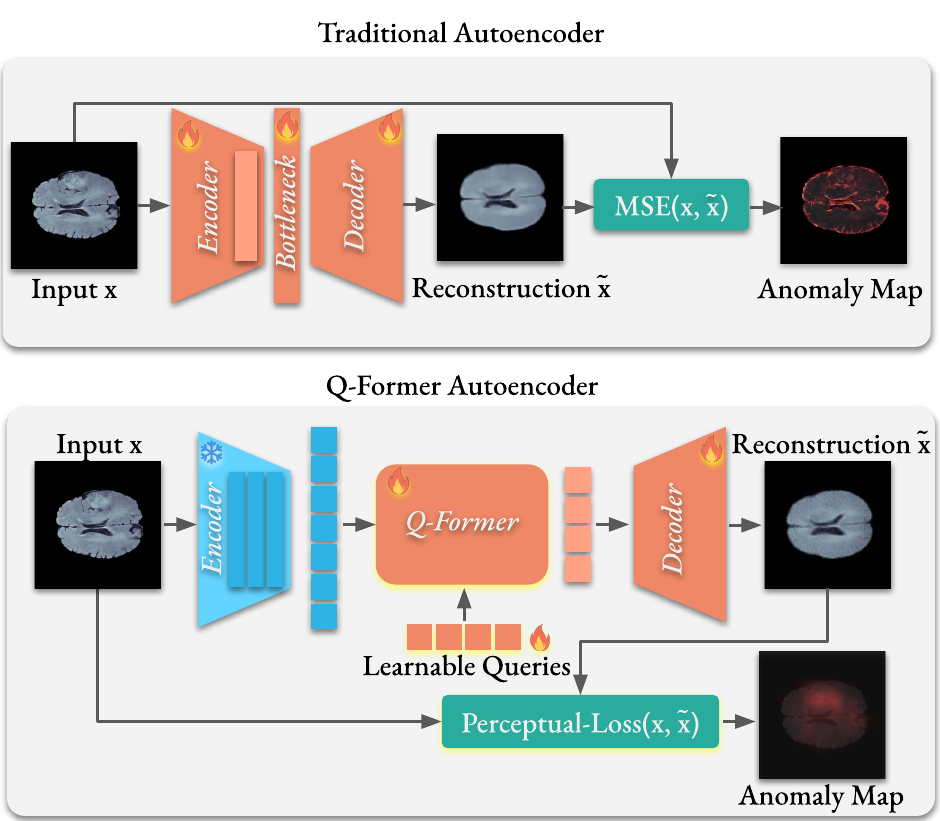}
      
  \caption{We illustrate the traditional autoencoder for anomaly detection (top) versus our Q-Former Autoencoder enhanced with Q-Former and perceptual loss (bottom).  The traditional autoencoder typically uses a trainable encoder-decoder pair and relies on Mean Squared Error (MSE) for optimization and anomaly detection. Our framework includes the following improvements (\textbf{\textcolor{Goldenrod}{highlighted in yellow})}: \textit{(i)}  \textbf{a frozen encoder} (employing powerful pretrained vision foundation models, such as DINO, DINOv2 and OpenCLIP), \textit{(ii)} \textbf{a Q-Former} acting as a \textit{dynamic, learnable bottleneck} for efficient representation, and \textit{(iii)}  the use of a \textbf{perceptual loss} function based on Masked Autoencoder. Our framework is able to produce meaningful anomaly detection \textit{precisely highlighting the anomalous regions} (bottom-right, \textbf{\textcolor{red}{in red}}).}  
  \label{fig:teaser}
  \vspace{-1.5em}
\end{figure}


Early work in unsupervised anomaly detection has predominantly relied on convolutional autoencoders trained to reconstruct normal images.
These conventional autoencoders suffered from limited representational power, restricting their effectiveness in anomaly detection. 
Recent advances in vision foundation models, such as DINO~\cite{caron_dino2021}, DINOv2~\cite{Oquab2023DINOv2LR}, and Masked Autoencoders (Masked AE)~\cite{He2021MaskedAA}, have demonstrated remarkable representation transferability to diverse tasks. Despite their potential, these models have been largely overlooked in the detection of medical image anomalies. One of the few exceptions is MVFA-AD~\cite{huang2022self} which employed the CLIP model~\cite{Radford2021LearningTV} to perform zero-shot and few-shot medical anomaly detection. Unfortunately, these methods often suffer a performance gaps compared to task-specific methods.

To bridge this gap, we propose a novel framework, \textbf{Q-Former Autoencoder}, that modernizes the autoencoder approach for unsupervised medical anomaly detection by integrating vision foundation models and an attention-based bottleneck mechanism based on Q-Former, as illustrated in Figure~\ref{fig:teaser}. 
First, we leverage pretrained vision foundation models, namely DINO~\cite{caron_dino2021}, DINOv2~\cite{Oquab2023DINOv2LR} and OpenCLIP~\cite{schuhmann2022laionb}, as frozen encoders, extracting robust and semantically rich features without requiring domain-specific retraining or fine-tuning. 
Second, we introduce a Q-Former model as a \textit{flexible bottleneck}, which aggregates multi-scale features and outputs a fixed-length latent representation. 
This design provides \textit{explicit control over the reconstruction granularity} while simultaneously improving the model's capacity to accurately represent normal structures. 
Third, we employ a perceptual loss computed using features extracted by a pretrained Masked Autoencoder, which  encourages reconstructions that preserve high-level semantics rather than low-level pixel details.

Our \textit{\textbf{modernized autoencoder}} significantly outperforms its standard counterpart in accurately detecting and localizing anomalies,  as illustrated in Figure~\ref{fig:teaser}. 
To evaluate our framework, we perform extensive experiments on four data sets from the BMAD~\cite{bmad} benchmark: BraTS2021~\cite{baid2021rsna, bakas2017advancing, menze2014multimodal}, RESC~\cite{hu2019automated}, RSNA~\cite{wang2017chestx}, and LiverCT~\cite{LiTs, landman2015Altas}. 

Our framework achieves state-of-the-art scores on all data sets reaching an AUROC of $94.3\%$ on BraTS2021 and $83.8\%$ on RSNA, showcasing its effectiveness across diverse image modalities, including MRI, OCT and X-rays.

In summary, our contributions are threefold: 

\begin{itemize}
\item We propose a \textbf{\textit{modernized and enhanced autoencoder}} approach that integrates frozen vision foundation models, a Q-Former bottleneck, and a perceptual loss for unsupervised anomaly detection. 

\item Our proposed framework achieves strong performance, reaching state-of-the-art AUROC scores on three medical anomaly detection benchmarks (namely BraTS2021, RESC, RSNA) without requiring domain-specific encoder finetuning. 

\item We provide detailed ablation experiments showing how vision foundation models, which are primarily trained on natural images, are able to generalize effectively to the medical image domain when combined with proper architectural adaptations. 
\end{itemize}


%% file: sec/2_related_work.tex
\section{Related Work}
\label{sec:formatting}

 \subsection{Taxonomy of Anomaly Detection Approaches}

\noindent
\textbf{Learning Strategy.} Image-based Anomaly Detection (AD) methods are commonly categorized into supervised, unsupervised, and zero-shot approaches.
Supervised methods, such as those based on few-shot learning or synthetic anomaly generation, require some access to annotated abnormal samples.  Zero-shot methods, on the other hand, aim to identify out-of-distribution samples without any access to domain data, often relying on pretrained models. Although promising in natural visual domains~\cite{Esmaeilpour_Liu_Robertson_Shu_2022}, they remain limited in specialized fields such as industrial inspection or medical imaging, where domain-specific knowledge is critical.
In such cases, unsupervised AD remains the most relevant setting. These methods train exclusively on normal samples and aim to detect deviations during inference. Although recent work has investigated multiclass AD~\cite{You_Cui_Shen_2022}, these approaches typically perform poorly compared to specialized algorithms, limiting their applicability in sensitive or safety-critical contexts such as medical analysis.

\begin{figure*}[t]
  \centering
  \includegraphics[width=0.99\linewidth]{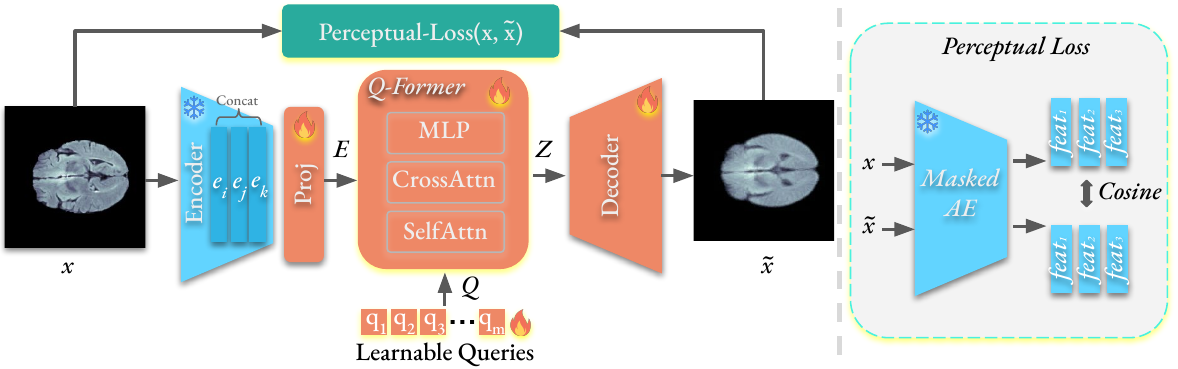}
      
  \caption{
  The training of our \textbf{Q-Former Autoencoder} for medical anomaly detection. Our framework uses \textit{a pretrained foundation model}, such as DINO~\cite{caron_dino2021}, DINOv2~\cite{Oquab2023DINOv2LR} or OpenCLIP~\cite{schuhmann2022laionb}, to extract \textit{multi-scale features} (\textit{E}).  These features, along with learnable query tokens (\textit{Q}), are processed by Q-Former acting as a \textit{dynamic bottleneck}. The output $z$ goes into the decoder to reconstruct $\Tilde{x}$. The \textit{Perceptual Loss} based on multi-scale features extracted from Masked AE~\cite{He2021MaskedAA} guides the training for semantic reconstruction.  }  
  \label{fig:method}
\end{figure*}

\noindent
\textbf{Feature-embedding or Reconstruction-based.} A classical taxonomy of AD methods~\cite{liu_deep_2024} divides them into two macro-categories: feature embedding and reconstruction-based approaches. Feature embedding methods rely on distances or density estimation in learned feature spaces. In contrast, reconstruction-based approaches, such as those based on autoencoders, learn to exclusively reconstruct normal data, assuming that anomalies cannot be effectively reconstructed from the model. These approaches have demonstrated strong performance, even when implemented as simple baselines with minimal architectural complexity~\cite{cai_medianomaly_2025}, while inherently supporting explainability and anomaly localization—particularly valuable in medical AD. In this work, we propose a framework based on reconstruction of the input.

\subsection{Autoencoder Architectures for AD}
Autoencoders learn a compressed latent representation of training data and attempt to re-project it to the input space. It is well-established that the compression of the latent representation is central to the anomaly detection (AD) capabilities of autoencoders~\cite{sakurada_anomaly_2014, cai_rethinking_2024}. The main challenge that such approaches face is to find a balance between a good reconstruction of normal images, while preventing the model to generalize to the anomalous samples.

A critical aspect is the choice of the reconstruction metric~\cite{pmlr-v172-meissen22a}: in addition to L2 loss, structural similarity index metric (SSIM) has been explored~\cite{bergmann_improving_2019, meissen_unsupervised_2023}, as well as perceptual losses~\cite{johnson_perceptual_2016, shvetsova_anomaly_2021}. Some of the most effective approaches recently proposed measure the distance in the feature space rather than the image space, showing robust results~\cite{meissen_unsupervised_2023, guo_encoder-decoder_2024, guo_recontrast_2023}. This approach is often used in combination with knowledge-distillation techniques, to further amplify the distance of anomalous samples~\cite{deng_anomaly_2022, tien_2023}.
Other relevant related methods involved the use of variational autoencoders~\cite{Marimont2020AnomalyDT}, masked autoencoders ~\cite{georgescu_masked_2023, xu_maediff_2024} and normalizing flow mechanisms~\cite{zhao_ae-flow_2022}. Similarly to the aforementioned works~\cite{johnson_perceptual_2016, shvetsova_anomaly_2021}, we employ the perceptual loss to train the autoencoder. However, different from the previous work, we utilize the Masked Autoencoder to guide the optimization of our model.

\subsection{Vision Foundation Models}
Recent advances in large-scale model pre-training have enabled the development of highly versatile foundation models for vision tasks, predominantly leveraging Vision Transformer (ViT)~\cite{Dosovitskiy2020AnII} architectures. Notable examples include CLIP~\cite{Radford2021LearningTV}, which employs a contrastive learning framework; DINOv2~\cite{Oquab2023DINOv2LR} and Masked Autoencoder~\cite{He2021MaskedAA} trained with various self-supervised schemes; and supervised models like SAM~\cite{Kirillov2023SegmentA}. These models, trained on large-scale data sets, learn rich representations that capture semantic and structural image information, enabling strong generalization across diverse downstream tasks.

The use of high-capacity vision foundation models for unsupervised anomaly detection (AD) remains underexplored. Zhang et al.~\cite{Zhang2025ExploringPV} established a multiclass AD baseline using frozen ViTs. More recent approaches leverage vision-language models: Jeong et al.~\cite{Jeong2023WinCLIPZA} employed compositional prompt ensembles and a sliding window for segmentation, and integrates memory banks to enable few-shot learning. Zhou et al.~\cite{Zhou2023AnomalyCLIPOP} used object-agnostic templates and prompt tuning. Huang et al.~\cite{Huang2024AdaptingVM} addressed domain shift through a dedicated adaptation module. Gu et al.~\cite{Gu2023AnomalyGPTDI} repurposed multimodal conversational models for AD, achieving strong results on industrial benchmarks.

While these models show promising performances, especially in zero- and few-shot regimes—a scalable, unified approach to fully leverage foundation models remains lacking, often resulting in performance gaps compared to task-specific methods. Therefore, in this work, we propose to fully leverage the foundation models, proposing an enhanced autoencoder framework equipped with Q-Former and Masked AE-based perceptual loss.   



%% file: sec/3_method.tex
\section{Q-Former Autoencoder}
\subsection{Overview} 
Autoencoder (AE) models are commonly employed in anomaly detection due to their ability to learn compact representations of normal data.
An autoencoder consists of an encoder, a latent space (bottleneck), and a decoder. 
The encoder compresses the input $x$ into a latent representation $z$, i.e., $z=\mathrm{Encoder}(x)$.
This latent representation is expected to capture the most informative aspects of the data. 
The decoder reconstructs the input from $z$, mapping it back to the input space, i.e.,  $\Tilde{x} = \mathrm{Decoder}(z)$. 
AEs are typically trained to minimize the reconstruction error between the input $x$ and its reconstruction $\Tilde{x}$.

Training autoencoders only on normal data enables anomaly detection, as the model should reconstruct normal data accurately while failing to reconstruct anomalies. 
 
\subsection{Evolving Autoencoders}
We present the training of our Q-Former Autoencoder (QFAE), highlighting the integration of Q-Former and Perceptual Loss in Figure~\ref{fig:method}.

\noindent
\textbf{Encoders.} Prior to the introduction of Vision Transformers (ViTs)~\cite{Dosovitskiy2020AnII}, Convolutional Neural Networks (CNNs) were the standard choice for encoders. 
Modern foundation models~\cite{caron_dino2021,He2021MaskedAA,Oquab2023DINOv2LR,Radford2021LearningTV}, however, mainly employ ViT architectures. 
ViTs begin by splitting the input image into non-overlapping patches and extracting patch representations using a shallow neural network. 
Subsequently, self-attention operation is applied on these patch representations, together with other normalization and feed-forward layers. These operations are repeated multiple times. 
As a result, the ViT encoder outputs a fixed-length sequence of patch embeddings.  

In this work, we employ pretrained encoders from foundation models, such as DINO~\cite{caron_dino2021}, DINOv2~\cite{Oquab2023DINOv2LR}, CLIP~\cite{Radford2021LearningTV}, Masked AE~\cite{He2021MaskedAA}.

\paragraph{Latent Space.} Inspired by BLIP-2~\cite{blip2} and BRAVE~\cite{brave}, we employ the Q-Former architecture as the bottleneck. 
Q-Former is well-suited because it processes variable-length contextual input to produce fixed-length latent codes, enabling the combination of tokens from different levels and even different architectures.  
The input to Q-Former is a set of learnable tokens, where the number of tokens controls the number of reconstructed patches. This enables the reconstruction of the output at multiple granularities (different patch sizes).
Q-Former interacts with the encoder features through a cross-attention layer.  
The encoder output serves as keys and values, being cross-attended by the Q-Former queries, as illustrated in Figure~\ref{fig:method}.
This design enables Q-Former to aggregate information from the latent features of the encoder efficiently, as Q-Former eliminates quadratic self-attention.
  
For a given input $x$, we obtain its embedding as $[e_i, e_j, \dots, e_k]$ 
where $[.]$ is the concatenation operation and $e_i, e_j, e_k$ are features from different layers of a pretrained ViT foundation model. 
These features are then adapted to the current task via a projection layer: $E=\mathrm{Proj}([e_i, e_j, \dots, e_k ])$, shown in Figure~\ref{fig:method}. 
We define the learnable queries $Q = [q_1, q_2, \dots, q_m]$, where $m$ is the desired length of the output sequence. 
A block of Q-Former is defined as: 

\begin{equation}
\begin{split} 
    & Q = \mathrm{SelfAttn}(Q) \\ 
    & Q = \mathrm{CrossAttn}(Q, E) \\ 
    & Z = \mathrm{MLP}(Q). \\ 
\end{split}
\end{equation}
\noindent  Based on our validation experiments, we employ only a single Q-Former block in our framework.

\paragraph{Decoder.} The decoder receives as input the latent representation $Z$ produced by Q-Former and reconstructs the original input image $x$. 
As noted earlier, the length of the reconstructed sequence is controlled by the number of learnable queries in Q-Former. 
The decoder architecture is a lightweight Transformer with only a few layers.  
Therefore, the reconstructed sequence of tokens $\Tilde{x}_{tok}=\mathrm{Decoder}(Z)$. In the last step, we reconstruct the input image by re-arranging the tokens $\Tilde{x}=\mathrm{unpatchify}(\Tilde{x}_{tok})$.

\paragraph{Perceptual Loss.}
Autoencoders are typically trained by minimizing the mean squared or mean absolute error between the input $x$ and its reconstruction $\Tilde{x}$.
In practice, perceptual loss has been proposed to improve reconstruction quality~\cite{johnson_perceptual_2016, shvetsova_anomaly_2021}. 
%
We compute perceptual loss using features extracted from different layers of a pretrained Masked Autoencoder (Masked AE)~\cite{He2021MaskedAA}. 
The perceptual loss minimizes the cosine distance between features from the original and reconstructed images: 

\begin{equation} 
  \mathcal{L}_{\text{Perceptual}} = \frac{1}{|I|}\sum_{i\in I}\left(1 - \frac{\mathrm{feat}_i \cdot \Tilde{\mathrm{feat}}_i}{||\mathrm{feat}_i||_2 ||\Tilde{\mathrm{feat}}_i||_2}\right)
\end{equation}

\noindent where $I$ is the set of selected layer indices from which features $\mathrm{feat}$ are obtained using the Masked AE~\cite{He2021MaskedAA}.

\paragraph{Anomaly Score Computation.}
We compute the anomaly score similarly to the perceptual loss, by comparing features extracted from multiple layers of the pretrained Masked AE, derived from the original and reconstructed images.

For each layer $i$ in a set of selected layers $I$, we extract the feature maps $\mathrm{feat}_i \in \mathbb{R}^{h_i \times w_i \times c}$ and $\Tilde{\mathrm{feat}}_i \in \mathbb{R}^{h_i \times w_i \times c}$, corresponding to the original input and its reconstruction, respectively. We then compute a layer-wise anomaly map, $A_{\text{map},i}$, by calculating the cosine distance at every spatial location $(j,k)$ between the corresponding feature vectors (patch embeddings).

\begin{equation}\label{eq:amap_intermediate}
  A_{\text{map}, i}(j,k) = 1 - \frac{\mathrm{feat}_{i, (j,k)} \cdot \Tilde{\mathrm{feat}}_{i, (j,k)}}{||\mathrm{feat}_{i, (j,k)}||_2 ||\Tilde{\mathrm{feat}}_{i, (j,k)}||_2}.
\end{equation}


\noindent
The final anomaly score for an image, a single scalar value, is calculated from these layer-wise maps by taking the maximum value from each map and averaging these maximums:

\begin{equation}\label{eq:score_reduction}
  A_{\text{score}} = \frac{1}{|I|}\sum_{i\in I}\max(A_{\text{map},i}).
\end{equation}

For visualization purposes, a consolidated anomaly map is generated by pixel-wise averaging all the layer-wise anomaly maps.

\begin{equation}\label{eq:amap_final}
  A_{\text{map, final}} = \frac{1}{|I|}\sum_{i\in I} A_{\text{map},i}.
\end{equation}

\noindent This final anomaly map enables the localization of anomalous regions within the image, as illustrated in Figure~\ref{fig:amaps}.

%% file: sec/4_results.tex
\section{Experiments}

\subsection{Data sets} We report results on three data sets: BraTS2021~\cite{baid2021rsna, bakas2017advancing, menze2014multimodal}, RESC~\cite{hu2019automated} and RSNA~\cite{wang2017chestx}, as detailed below. 
Additional results on the LiverCT~\cite{LiTs, landman2015Altas} data set are provided in the supplementary material.

\noindent
\textbf{BraTS2021.} The BraTS2021~\cite{baid2021rsna, bakas2017advancing, menze2014multimodal} data set, part of the the BMAD~\cite{bmad} benchmark, contains brain MRI images with pixel-level annotations of various anomalies.
BraTS2021 has a total of $11,\!298$ images, split into $7,\!500$ training, $83$ validation and $3,\!715$ test samples. 
Each slice has a resolution of $240 \times240$ pixels.

\noindent
\textbf{RESC.}  RESC~\cite{hu2019automated}, also part of BMAD~\cite{bmad}, contains retinal OCT images. 
The data includes $6,\!217$ images in total, with $1,\!805$ used for testing. 
All images are high-resolution with the size of $512 \times 1024$ pixels.

\noindent
\textbf{RSNA.} The RSNA data set~\cite{wang2017chestx}, included in the BMAD~\cite{bmad} benchmark, consists of chest X-ray images with image-level anomaly annotations.
It contains $26,\!684$ images of resolution $1024\times1024$, split into   $8,\!000$ training, $1,\!490$ validation and $17,\!194$ test samples.

\subsection{Implementation Details}
We employed different pretrained vision foundation models as encoders, including DINO~\cite{caron_dino2021}, DINOv2~\cite{Oquab2023DINOv2LR}, OpenCLIP~\cite{schuhmann2022laionb} and Masked Autoencoder~\cite{He2021MaskedAA}. 
As previously stated, the encoder remains frozen during the training of the framework. 
Our decoder is a Transformer architecture, consistent with the Masked AE setting, with $6$ layers, $12$ heads, and a hidden dimension of $768$. 
Features are extracted from layers 20 and 22 of the ViT-L encoder and layers 8 and 10 of the ViT-B architecture. 
The architecture of Q-Former consists of only one Transformer layer. 
The number of learnable tokens in Q-Former is determined by the reconstruction patch size. 
With a patch size of $8\times8$ pixels and an input resolution of $224\times224$, the number of learnable tokens is $784$ (i.e.: $784 = (224 / 8) ^ 2$). 
Both the Q-Former and the decoder are trained for 300 epochs using perceptual loss.
Hyperparameters were tuned on the validation sets. 
Further implementation details are presented in the supplementary material.

\noindent
\textbf{Evaluation Metrics.} Consistent with previous work~\cite{bmad,Huang2024AdaptingVM}, we report the Area Under the Receiver Operating Characteristic (AUROC) curve for anomaly detection. 
AUROC for localization is not reported due to its tendency to produce overly optimistic scores in cases of severe pixel-wise class imbalance, which is prevalent in anomaly detection.

\begin{table}[t]
    \centering
    \setlength\tabcolsep{2.5pt}
    \caption{Ablation results on BraTS2021~\cite{baid2021rsna, bakas2017advancing, menze2014multimodal} of our medical anomaly detection framework, QFAE. We demonstrate step by step  how incorporating components, such as Q-Former and perceptual loss, elevates a simple AE model to a strong medical anomaly detector using off-the-shelf models. MAE: Mean Absolute Error. $\mathcal{L}_{\text{Perceptual}}$: Perceptual loss based on the specified model.}
    \scalebox{0.99}{
    \begin{tabular}{lcccc}
    \toprule
    & \bf Q-Former   &  \bf  Loss   & \bf AUROC (\%)\\
    \hline
    \rownumber{1}&  \xmark  & MAE  &  $66.6$\\
    \rownumber{2}&  \cmark  & MAE  &   $79.5$ \\  
    \rownumber{3}& \cmark   & $\mathcal{L}_{\text{Perceptual}}$ (Masked ViT) & $86.8$   \\

    \bottomrule
    \end{tabular}
    }

    \label{tab:new_architecture} 
\end{table}

\begin{table*}[t] 
\caption{Ablations results on the BraTS2021~\cite{baid2021rsna, bakas2017advancing, menze2014multimodal} data set changing different components of our architecture. Perceptual loss achieves higher performance than the simple mean absolute error (MAE) optimization, along with taking the maximum of the error. We also notice that using multiple hidden layers from the perceptual encoder is better along with using a smaller patch size for the decoder. 
The default configuration is highlighted in {\color{CornflowerBlue}{light blue}}. $\text{L}_{\text{Perceptual}}$: Perceptual loss based on Masked AE. }
\label{tab:ablation}
\centering
\subfloat[
\textbf{Loss function}. Mean Absolute Error (MAE) decreases the performance when combined with the perceptual loss. The top performance is obtained with  $\text{L}_{\text{Perceptual}}$.
\label{tab:ablationbeta}
]{
\centering
\begin{minipage}{0.28\linewidth} \label{tab:ablation_losses}
{\begin{center} 
\begin{tabular}{lc}
Loss & AUROC \\
\hline 
MAE & 79.0 \\
MAE, $\text{L}_{\text{Perceptual}}$ & $79.2$ \\
\rc $\text{L}_{\text{Perceptual}}$ & \rc $88.5$ \\
\end{tabular}
\end{center}}\end{minipage}
} 
\hspace{1em}
\subfloat[
\textbf{Aggregation in Eq.~\ref{eq:score_reduction}.} Selecting the maximum error within Eq.~\ref{eq:score_reduction} yields top performance.  
\label{tab:ablation_reduction}
]{
\centering
\begin{minipage}{0.19\linewidth}{\begin{center}
\begin{tabular}{lc}
Function & AUROC\\
\hline
mean & $88.5$ \\
\rc max & \rc $92.6$ \\
\end{tabular}
\end{center}}\end{minipage}
} 
\hspace{1em}
%
\subfloat[
\textbf{Layers from the perceptual model}. Using multiple layers from the perceptual model achives top performance.
\label{tab:ablation_perceptual}
]{
\centering
\begin{minipage}{0.2\linewidth}{\begin{center}
\begin{tabular}{lc}
Layers & AUROC \\
\hline
5, 11  & $92.6$ \\
\rc 11, 15, 19 & \rc $93.0$ \\

\end{tabular}
\end{center}}\end{minipage}
} 
\hspace{1em}
\subfloat[
\textbf{Decoder patch size}. Reconstructing the input using smaller patch sizes achieves top performance.
\label{tab:ablation_decoder_patch_size}
]{
\centering
\begin{minipage}{0.2\linewidth}{\begin{center}
\begin{tabular}{lc}
Patch size & AUROC\\
\hline
\rc 8 &  \rc $93.0$  \\
16 &  $92.5$  \\
32 &  $91.1$   \\ 
\end{tabular}
\end{center}}\end{minipage}
}

\end{table*}

\subsection{Ablation Study}
\noindent  
\textbf{Establishing the New Architecture.}
We ablate each component of our Q-Former Autoencoder and present the results on the BraTS2021~\cite{baid2021rsna, bakas2017advancing, menze2014multimodal} data set in Table~\ref{tab:new_architecture}. 
To create an updated autoencoder architecture, we start with the basics of employing a pretrained encoder and training a decoder (row \rownumber{1}). 
We selected DINOv2 ViT-B/14 as the encoder due to its exceptional results on zero-shot tasks. 
The AE architecture contains only the encoder and the decoder without any bottleneck introduced, and it was trained by minimizing the mean absolute error between the input and the output of the decoder.  
This basic version of AE reaches an AUCROC score of only $66.6$. 
Adding the Q-Former module as the bottleneck (row \rownumber{2}) improves the AUROC by $12.9$ (from $66.6$ to $79.5$) showing that Q-Former is capable of retaining the structure of normal data, which makes it a good choice for anomaly detection.  
Lastly, changing the optimization loss from the mean absolute error to the perceptual loss computed based on Masked AE features, increases the performance to $86.8$ (row \rownumber{3}). 
By applying these designed choices (Q-Former, perceptual loss), we evolve a simple AE architecture with modest results to a powerful and accurate framework that reaches strong performances.

\noindent  
\textbf{The Impact of the Loss Function.} We evaluate the impact of the loss function on the detection performance for medical anomalies, reporting the results in Table~\ref{tab:ablation_losses}. 
Using the mean absolute error alone or even combining it with the perceptual loss produces poor results. 
Training solely with the perceptual loss ($\mathcal{L}_{\text{Perceptual}}$) achieves the best performance, highlighting the superiority of deep feature-based optimization over pixel-level reconstruction. 

\noindent  
\textbf{The Impact of the Aggregation.} We further evaluate the influence of different aggregation strategies on the anomaly score computation, with results reported in Table~\ref{tab:ablation_reduction}. 
Defining the anomaly score as the maximum reconstruction error produces the best performance, which aligns with the intuition that anomalies are inherently harder to reconstruct.

\noindent  
\textbf{The Impact of Perceptual Features.} We analyze the effect of perceptual features when extracted from different layers of the Masked AE model~\cite{He2021MaskedAA}, reporting the results in Table~\ref{tab:ablation_perceptual}. 
In our initial experiment, we extracted features from layers 5 and 11 to guide model optimization, achieving a performance of $92.6$. 
Adding another layer further improves the performance to $93.0$, demonstrating that incorporating additional signals during AE training is beneficial for robust anomaly detection.

\noindent  
\textbf{The Impact of the Decoder Patch Size.} Employing the Q-Former architecture as the bottleneck decouples the dependency between encoder and decoder output lengths. 
Therefore, the decoder can reconstruct the input at varying granularities (different patch sizes). 
We evaluated different patch sizes such as $8\times8$, $16\times16$ and $32\times32$, reporting the results in Table~\ref{tab:ablation_decoder_patch_size}. 
As anticipated, smaller patch sizes produce higher performance, enabling the decoder to generate more precise reconstructions.

\noindent  
\textbf{Impact of Perceptual Model Patch Size.} Feature extraction for computing the perceptual loss is entirely independent of the framework's encoder and decoder, which enables the use of multi-scale patch sizes to compute the perceptual features. 
Interestingly, Table~\ref{tab:patch_size_vitmae} reveals that larger patches lead to improved anomaly detection performance. 
However, the best performance of $94.4$ AUROC on BraTS2021 is achieved by combining two large patch sizes ($32\times32$ and $56\times56$ pixels), effectively creating a pyramid of features. 
This finding suggests that larger patch sizes better capture the structure of the data, making it easier to spot the differences, thus improving anomaly detection.   

\begin{table}
  \centering
  \caption{Anomaly detection results on BraTS2021~\cite{baid2021rsna, bakas2017advancing, menze2014multimodal} in terms of AUROC (\%) when changing the patch size of the Masked AE~\cite{He2021MaskedAA}. We notice that dividing the input into larger patches significantly improves the performance. Top results are highlighted in bold. The default configuration is highlighted in {\color{CornflowerBlue}{light blue}}.}
  \begin{tabular}{@{}lc@{}}
    \toprule
    \bf Masked AE Input patch size { }{ }{ }{ }{ }{ } { }{ }{ }{ }{ }{ } { }{ }{ }{ }{ }{ } &  \bf AUROC \\
    \midrule
    
    16 & $72.7$ \\
    56 & $92.8$ \\
    
    16, 32, 56 & $93.0$ \\
    \rc 32, 56  & \rc $\mathbf{94.4}$ \\

    \bottomrule
  \end{tabular}
  
  \label{tab:patch_size_vitmae}
\end{table}

\begin{table}
  \centering
   \caption{Anomaly detection results on BraTS2021~\cite{baid2021rsna, bakas2017advancing, menze2014multimodal} in terms of AUROC (\%) when different pretrained encoders are employed. Notably, Masked AE~\cite{He2021MaskedAA} encoder obtains poor performance due to its ability to reconstruct the input. Both DINO~\cite{caron_dino2021} and DINOv2~\cite{Oquab2023DINOv2LR} achieve strong performance. The default configuration is highlighted in {\color{CornflowerBlue}{light blue}}.}
  \begin{tabular}{@{}lc@{}}
    \toprule
     \bf Encoders &  \bf AUROC  \\
    \midrule
    DINO ViT-B/8 & $94.3$  \\
    OpenCLIP ViT-L/14 &  $94.0$  \\
    Masked AE ViT-L/16 &  $71.5$  \\
    \rc DINOv2 ViT-L/14 & \rc $94.4$  \\

    DINOv2 ViT-L/14 + DINO ViT-B/8 & $94.5$ \\

    DINOv2 ViT-L/14 + OpenCLIP ViT-L/14  & $93.6$ \\

    DINOv2 ViT-L/14 + OpenCLIP ViT-B/32 & $94.3$  \\

    DINOv2 ViT-L/14 + Masked AE ViT-B/16  & $76.7$ \\

    DINOv2 ViT-L/14 + Masked AE ViT-L/16   & $74.3$\\

    \bottomrule
  \end{tabular}
 
  \label{tab:diff_encoders}
\end{table}

\begin{table}[t]
    \centering
    \renewcommand{\arraystretch}{1.5}  
       
    \caption{Anomaly detection performance (mean + std) on BraTS2021~\cite{baid2021rsna, bakas2017advancing, menze2014multimodal}, RESC~\cite{hu2019automated} and RSNA~\cite{wang2017chestx}. The results are reported for five repetitions of the experiment. $^*$:~denotes only three repetitions. The top results are reported in bold. Our method is able to outperform all methods obtaining state-of-the-art performance on all three data sets.
    }
    \resizebox{0.99\linewidth}{!}{
    \begin{tabular}{l|ccc}
        \toprule
        \bf  Methods & \bf BraTS2021 & \bf RESC &  \bf RSNA \\ 
        \midrule
         f-AnoGAN~\citep{schlegl2019f} & $77.3 \pm 0.18$  & $77.4 \pm 0.85$  &  $55.6 \pm 0.09$  \\
         GANomaly~\citep{akcay2018ganomaly}& $74.8 \pm 1.93$ & $52.6 \pm 3.95$ & $62.9\pm 0.65$  \\
         DRAEM~\citep{zavrtanik2021draem} & $62.4 \pm 9.03$ & $83.2 \pm 8.21$ & $67.7 \pm 1.72$   \\
             UTRAD~\citep{chen2022utrad}  & $82.9 \pm 2.32$  & $89.4 \pm 1.92$ & $75.6 \pm 1.24$ \\ 
         
        DeepSVDD~\citep{ruff2018deep}  & $87.0 \pm 0.66$ & $74.2 \pm 1.29$  & $64.5 \pm 3.17$ \\
             CutPaste~\citep{li2021cutpaste}  & $78.8 \pm 0.67$  & $90.2 \pm 0.61$  & $82.6 \pm 1.22$ \\
             SimpleNet~\citep{liu2023simplenet} & $82.5 \pm 3.34$ & $76.2 \pm7.46$ & $69.1 \pm 1.27$ \\
             
             MKD~\citep{salehi2021multiresolution} & $81.5 \pm 0.36$  &  $89.0 \pm 0.25$  &  $82.0 \pm 0.12$ \\
             
             RD4AD~\citep{deng2022anomaly} &  $89.5 \pm 0.91$ &$87.8 \pm 0.87$ &  $67.6 \pm 1.11$ \\
            
             STFPM~\citep{yamada2021reconstruction}  & $83.0 \pm 0.67$ &  $84.8 \pm 0.50$  & $72.9 \pm 1.96$ \\
             PaDiM ~\citep{defard2021padim} & $ 79.0 \pm 0.38 $& $75.9 \pm0.54$  & $77.5 \pm 1.87$  \\
             PatchCore~\citep{roth2022towards}  & $91.7 \pm 0.36$  &  $91.6 \pm0.10$  & $76.1 \pm 0.67$\\
             CFA~\citep{lee2022cfa}  & $84.4 \pm 0.87$ & $69.9 \pm 0.26$ &  $66.8 \pm 0.23$ \\
            CFLOW~\citep{gudovskiy2022cflow} & $74.8 \pm 5.32$  & $75.0 \pm 5.81$  & $71.5 \pm 1.49$    \\
             CS-Flow~\citep{rudolph2022fully}  & $90.9 \pm 0.83$ &  $87.3 \pm 0.58$ & $83.2 \pm 0.46$  \\ 
             P-VQ$^*$~\citep{pvq}  &  $94.3 \pm 0.23$ &  $89.0 \pm 0.48$ & $79.2 \pm 0.04$  \\ 
              \rc \bf QFAE (ours)  & \rc $\mathbf{94.3 \pm 0.18}$ & \rc   $\mathbf{91.8 \pm 0.55}$ & \rc   $\mathbf{83.8 \pm 0.46}$  \\ 
        \bottomrule
    \end{tabular}
  }
    \label{tab:main_results}
    \vspace{-1em}
\end{table}

\noindent  
\textbf{Impact of the Encoder.} We evaluate different combination of encoders including DINO~\cite{caron_dino2021}, DINOv2~\cite{Oquab2023DINOv2LR}, OpenCLIP~\cite{schuhmann2022laionb} and Masked AE~\cite{He2021MaskedAA} reporting the anomaly detection results on BraTS2021 in Table~\ref{tab:diff_encoders}.
Among the single encoders, DINOv2~\cite{Oquab2023DINOv2LR} demonstrated the best performance of $94.4$ AUROC, underscoring its strong capability on zero-shot tasks. 
When combining DINOv2~\cite{Oquab2023DINOv2LR} with DINO~\cite{caron_dino2021}, the performance slightly improves reaching an AUROC of $94.5$.
However, we concluded that this improvement does not justify the computational burden of adding an extra encoder. 
Therefore, DINOv2~\cite{Oquab2023DINOv2LR} was selected as the default single encoder for our framework. 
Notably, the Masked AE~\cite{He2021MaskedAA} encoder exhibits poor performance, even when combined with DINO~\cite{caron_dino2021} or DINOv2~\cite{Oquab2023DINOv2LR}, primarily due to its strong reconstructive capacity that hinders anomaly discrimination. 

The results from DINO~\cite{caron_dino2021}, DINOv2~\cite{Oquab2023DINOv2LR}, and OpenCLIP~\cite{schuhmann2022laionb} demonstrate that foundation models are effective for anomaly detection, even within the medical domain.

\begin{figure}[t]
  \centering
  \includegraphics[width=0.99\linewidth]{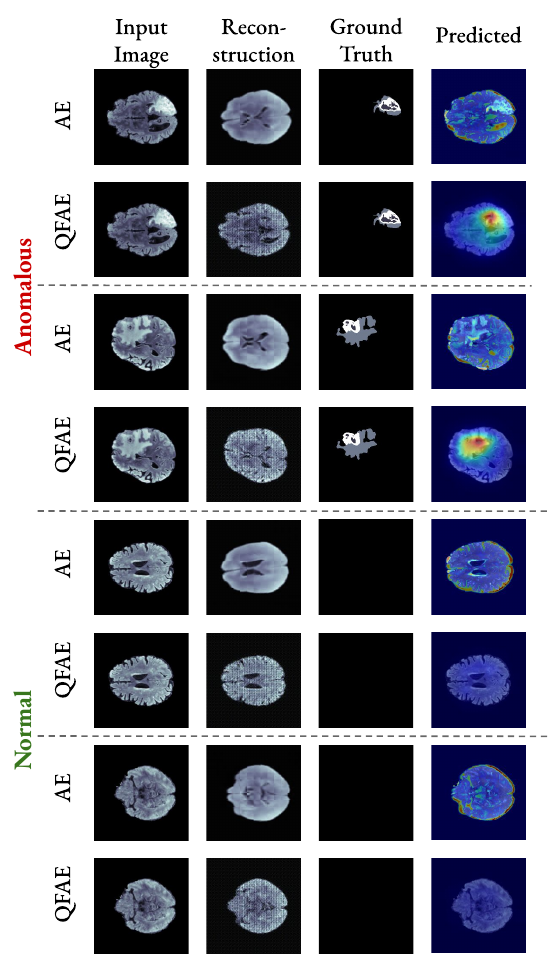}
      
  \caption{Qualitative examples of anomaly localization on several samples from the BraTS2021~\cite{baid2021rsna, bakas2017advancing, menze2014multimodal} data set. For each sample, we present the original input, the reconstruction, the ground-truth and the predicted anomaly map. Both normal and abnormal samples are presented. Our Q-Former AutoEncoder (QFAE) with a traditional AutoEncoder (AE). Notably, our QFAE method consistently produces sharper and more accurate anomaly localizations compared to the baseline, closely aligning with the ground~truth. Moreover, our QFAE predicts very low anomaly scores for normal samples, being able to correctly identify them as normal samples.}  
  \label{fig:mae_vs_perceptual_amaps}
  \vspace{-2em}
\end{figure}

\begin{figure*}[t]
  \centering
  \includegraphics[width=0.99\linewidth]{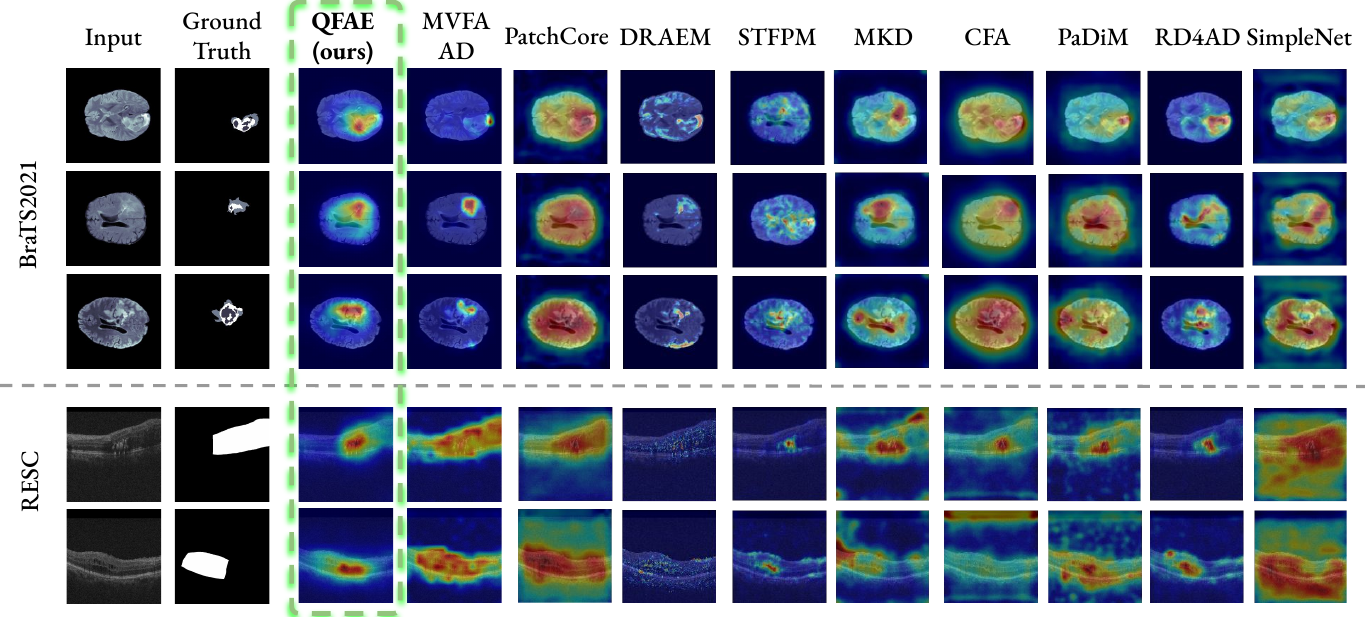}
      
  \caption{
  Qualitative examples of anomaly localization on several samples from the BraTS2021~\cite{baid2021rsna, bakas2017advancing, menze2014multimodal} and RESC~\cite{hu2019automated} data sets. For each sample, the columns show the original input, ground truth anomaly masks, and anomaly maps predicted by QFAE (ours) alongside various baseline methods. \textit{We note that MVFA-AD~\cite{Huang2024AdaptingVM} uses a few-shot strategy, therefore it is not unsupervised as the rest of the methods including ours}. The predicted anomalies for the baselines are cropped directly from BMAD~\cite{bmad}. Notably, our QFAE method consistently produces sharper and more accurate anomaly localizations compared to other approaches, closely aligning with the ground~truth.
  }  
  \vspace{-1.5em}
  \label{fig:amaps}
\end{figure*}

\subsection{Comparison with State-of-the-Art}
We compare our framework QFAE against several state-of-the-art methods on BraTS2021~\cite{baid2021rsna, bakas2017advancing, menze2014multimodal}, RESC~\cite{hu2019automated} and RSNA~\cite{wang2017chestx}, presenting the results in Table~\ref{tab:main_results}. 
We report the mean and standard deviation (std) of the results obtained from 5 independent runs for each experiment.

Our method achieves state-of-the-art performance on all data sets. 
In particular, on the BraTS2021 data set, our framework achieves an AUROC of $94.3 \pm 0.18$, on par with the previous best performing method ($94.3 \pm 0.23$ achieved by P-VQ~\cite{pvq}). 
This result demonstrates strong anomaly detection capabilities in brain imaging obtained simply by enhancing the standard AE framework.

Furthermore, our method outperforms all baselines on all three data sets. 
On RESC, we achieve the highest AUROC score of $91.8 \pm 0.55$, surpassing the previous state-of-the-art results of PatchCore~\cite{roth2022towards} ($91.6 \pm 0.10$). 
On RSNA, QFAE achieves an AUROC score of $83.8 \pm 0.46$, outperforming the next best method, CS-Flow~\cite{rudolph2022fully} ($83.2 \pm 0.46$).

These top results highlight the robustness of our framework across several medical imaging modalities (MRI, X-rays, and OCT). 
Additionally, this work further highlights that foundation models, primary trained on natural images, can be successfully employed in a different domain, such as medical images, without additional finetuning.

\subsection{Qualitative Results} 
We present qualitative results in Figure~\ref{fig:mae_vs_perceptual_amaps} and Figure~\ref{fig:amaps}. We illustrate samples from the BraTS2021~\cite{baid2021rsna, bakas2017advancing, menze2014multimodal} and RESC~\cite{hu2019automated} data sets, along with the input image, the ground-truth anomaly mask, the anomaly map predicted by several state-of-the-art methods and our QFAE framework in Figure~\ref{fig:amaps}. On both data sets, our framework precisely localizes the anomalies. Notably, on the second BraTS2021 sample, anomaly localization proved challenging for most methods, with only MVFA-AD~\cite{Huang2024AdaptingVM} (\textit{few-shot}), DRAEM~\cite{zavrtanik2021draem}, and our enhanced AE (QFAE) achieving correct localization among the 10 evaluated approaches. This highlights that our approach is capable to accurately identify subtle and difficult-to-detect anomalies.

Additionally, in Figure~\ref{fig:mae_vs_perceptual_amaps}, we illustrate qualitative results comparing the anomaly results of our QFAE with those obtained by a traditional AE. The traditional AE employed a pretrained encoder and a decoder. We observed that our enhanced AE predicts anomalies that correlate well with the ground truth, while also yielding very low anomaly scores for normal samples. These results clearly show that using the Q-Former as a bottleneck is effective in detecting and localizing anomalies. Additionally, these findings indicate that the combined approach of incorporating Q-Former as a bottleneck and leveraging the Masked AE for perceptual loss is highly effective in enhancing medical anomaly detection performance.

%% file: sec/5_conclusion.tex
\section{Conclusions} 
In this paper, we introduced the \textbf{Q-Former AutoEncoder} (QFAE), a modernized autoencoder framework that leverages the power of state-of-the-art pretrained vision foundation models for medical anomaly detection. Our framework addresses key limitations of traditional autoencoders by integrating frozen pretrained encoders (DINO~\cite{caron_dino2021}, DINOv2~\cite{Oquab2023DINOv2LR} and OpenCLIP~\cite{schuhmann2022laionb}) for robust feature extraction, employing a trainable Q-Former as a \textit{dynamic bottleneck to produce fixed-length latent codes out of variable-length contextual input}, and utilizing a perceptual loss function for semantically meaningful reconstruction.  We rigorously evaluated QFAE on four diverse medical anomaly detection benchmarks: BraTS2021, RESC, RSNA, and LiverCT. Our results consistently \textit{demonstrate state-of-the-art performance across these data sets}, achieving superior AUROC scores and precise anomaly localization. Our work highlights the successful and robust application of large-scale, pretrained vision foundation models (initially trained on natural images) for unsupervised anomaly detection in specialized medical imaging domains, notably without requiring extensive fine-tuning. In future work, we plan to apply QFAE to multi-class medical anomaly detection.
\noindent
\textbf{Limitations.} Despite its strong performance, our proposed QFAE framework has certain limitations. While using pretrained foundation models, such as DINO, DINOv2, Masked AE, etc., enhances the generalization capabilities and reduces training time, it inherently limits the model's ability to learn domain-specific features. Despite our framework achieving consistently good results across modalities and data sets, we cannot claim that it will generalize to all anomaly types or varying levels of input complexity.

%% file: sec/supp.tex
We provide additional implementation details in Section~\ref{subsec:implementation_details}, additional experiments on LiverCT and RSNA in Section~\ref{subsec:liver_supp} and Section~\ref{subsec:chest_supp}.

\section{Implementation Details}
\label{subsec:implementation_details}

This section provides an overview of the implementation details for our proposed framework, ensuring full reproducibility of our results. All experiments were conducted in PyTorch.

\subsection{Hyperparameters}
The main hyperparameters used for training and evaluation are detailed in Table~\ref{tab:train_hyperparams} and Table~\ref{tab:eval_hyperparams}, respectively.

\begin{table}[h!]
\centering
\caption{Training hyperparameters for the experiments.}
\label{tab:train_hyperparams}
\resizebox{\columnwidth}{!}{%
\begin{tabular}{@{}lll@{}}
\toprule
\textbf{Component} & \textbf{Parameter} & \textbf{Value} \\ \midrule
\textbf{General} & Seed & 42, 7, 13, 65, 91 (mean of 5 runs are reported)\\
 & Image Resolution (Resize) & 224x224 \\
 & Batch Size & 64 \\
 & Epochs & 300 \\
 & Device & CUDA \\ \midrule
\textbf{Encoder} & Pre-trained Model & ViT-Large (ViT-L/14) with register tokens \\
 & Pre-training Method & DINOv2 \\
 & Frozen During Training & True \\
 & Hidden States Used & Features from the 2nd and 4th to last blocks \\
 & Final Projection In-Features & 1024 \\
 & Final Projection Out-Features & 768 \\ \midrule
\textbf{Q-Former (Junction)} & Number of Transformer Blocks & 1 \\
 & Internal Dimension & 768 \\
 & Output Dimension & 768 \\
 & Number of Learnable Queries & 784 (for 28x28 output patches) \\
 & Attention Heads & 8 \\
 & MLP Expansion Ratio & 4.0 \\ \midrule
\textbf{Decoder} & Internal Dimension & 768 \\
 & Depth (Number of Layers) & 6 \\
 & Attention Heads & 12 \\
 & Output Patch Size & 8x8 \\
 & Number of Output Patches & 28x28 \\
 & MLP Expansion Ratio & 4.0 \\ \midrule
\textbf{Optimization} & Optimizer & Adam \\
 & Learning Rate (Maximum) & $8 \times  10^{-5}$ \\
 & Learning Rate Scheduler & OneCycleLR \\ \midrule
\multirow{4}{*}{\textbf{Perceptual Loss}} & Pre-trained Perceptual Model & Masked Autoencoder (MAE) with ViT-Large Encoder \\
 & Distance Metric & Cosine Distance \\
 & Layers Used for Feature Extraction & From the 16th and 20th transformer blocks \\
 & Multi-Scale Input Patch Sizes & 32x32, 56x56 \\ \bottomrule
\end{tabular}%
}
\end{table}

\begin{table}[h!]
\centering
\caption{Evaluation configuration for the experiments.}
\label{tab:eval_hyperparams}
\resizebox{\columnwidth}{!}{%
\begin{tabular}{@{}lll@{}}
\toprule
\textbf{Component} & \textbf{Parameter} & \textbf{Value} \\ \midrule
\textbf{General} & Batch Size & 64 \\
 & Test Data Augmentation & None (only resize and normalize) \\ \midrule
\multirow{4}{*}{\textbf{Perceptual Metric}} & Pre-trained Perceptual Model & MAE with ViT-Large Encoder \\
 & Distance Metric & Cosine Distance \\
 & Layers Used for Feature Extraction & From the 12th, 16th, and 20th transformer blocks \\
 & Multi-Scale Input Patch Sizes & 16x16, 32x32, 56x56 \\ \midrule
\multirow{2}{*}{\textbf{Image-Level Score Aggregation}} & Spatial Aggregation per Feature Map & Max \\
 & Cross-Feature Map Aggregation & Mean \\ \midrule
\multirow{1}{*}{\textbf{Pixel-Level Map Aggregation}} & Cross-Feature Map Aggregation & Mean \\ \bottomrule
\end{tabular}%
}
\end{table}

\subsection{Perceptual Loss Formulation}
The training objective is to minimize a multi-scale perceptual loss. This loss is calculated in a three-step process:

\textbf{Step 1: Feature Extraction.}
For an input image $x$ and its reconstruction $\tilde{x}$, we extract feature maps from a set of pretrained perceptual models. We use multiple Masked Autoencoder (Masked AE) models, each distinguished by its input patch size $p \in P$. For each model, we select features from a set of transformer blocks $i \in I$. Let $\Phi_{i,p}(x)$ be the feature map of shape $C_i \times H_i \times W_i$ extracted from the $i$-th layer of the perceptual model with patch size $p$.

\textbf{Step 2: Anomaly Map Calculation.}
For each selected feature map, we compute an intermediate anomaly map, $A_{i,p}$, by calculating the cosine distance between the features of the original image and its reconstruction at every spatial location $(j,k)$.
\[
A_{i,p}(j,k) = 1 - \frac{\Phi_{i,p}(x)_{j,k} \cdot \Phi_{i,p}(\tilde{x})_{j,k}}{\|\Phi_{i,p}(x)_{j,k}\|_2 \cdot \|\Phi_{i,p}(\tilde{x})_{j,k}\|_2}
\]
This produces a set of single-channel anomaly maps, one for each combination of layer $i$ and patch size $p$.

\textbf{Step 3: Hierarchical Aggregation and Final Loss.}
The final loss is computed using a two-stage hierarchical aggregation. First, for each feature layer $i \in I$, we create a robust, layer-specific anomaly map, $A_{\text{combined}, i}$, by performing an element-wise multiplication of its corresponding anomaly maps from all different patch-size models $p \in P$. This step enforces a strict consensus across multiple scales for each feature level.
\[
A_{\text{combined}, i} = \prod_{p \in P} \text{Resize}_{(H,W)}(A_{i,p})
\]
Second, the total loss $\mathcal{L}$ is calculated by averaging the mean value of each of these robust, layer-specific maps. This treats the error signal from each feature layer as an independent contribution to the total loss.
\[
\mathcal{L}(x, \tilde{x}) = \frac{1}{|I|}\sum_{i \in I} \text{mean} \left( A_{\text{combined}, i} \right)
\]
For training, we use patch sizes $P=\{32, 56\}$ and features from the 16th and 20th transformer blocks of the Masked AE ViT-Large encoder.

\subsection{Anomaly Score and Map Generation}
During evaluation, we generate both an image-level scalar score for AUROC computation and a pixel-level anomaly map. Both start from the same set of intermediate anomaly maps, $A_{i,p}$, though computed using the evaluation configuration (Table \ref{tab:eval_hyperparams}). Let this evaluation set of maps be denoted by $\mathcal{A} = \{A_1, A_2, ..., A_N\}$.

\paragraph{Image-Level Anomaly Score Aggregation.}
To derive a single scalar score for each image, we perform a two-step aggregation:

\noindent
\textbf{Step 1: Spatial Aggregation.} For each anomaly map $A_n \in \mathcal{A}$, we find the maximum pixel value. This value, $s_n$, represents the most severe reconstruction error detected by that specific feature map.
\[
s_n = \max_{j,k} (A_n(j,k))
\]
\textbf{Step 2: Cross-Feature Aggregation.} The final image-level score, $A_{\text{score}}$, is the mean of these maximum values, averaged over all $N$ feature maps.
\[
A_{\text{score}} = \frac{1}{N} \sum_{n=1}^{N} s_n
\]
This method gives a robust score that is sensitive to strong local anomalies while benefiting from the diversity of features from different layers.

\paragraph{Pixel-Level Anomaly Map Generation.}
To generate a final 2D anomaly map, we use a different aggregation strategy that preserves spatial information. At each spatial location $(j,k)$, we take the mean value across all $N$ resized anomaly maps.
\[
A_{\text{pixel-max}}(j,k) = \max_{n \in \{1..N\}} (A_n(j,k))
\]

\subsection{Training and Data Augmentation}

The model is trained using the Adam optimizer with a \texttt{OneCycleLR} learning rate scheduler. To encourage the model to learn robust and generalizable representations of normal data, the following data augmentations are applied to the training set:
\begin{itemize}
    \item \textbf{Random Resized Crop:} Images are cropped to a random size (90\% to 100\% of the original) and aspect ratio (80\% to 120\% of the original) before being resized to the final input dimension.
    \item \textbf{Random Rotation:} Images are rotated by a random angle between -10 and +10 degrees.
    \item \textbf{Random Vertical Flip:} Images are flipped vertically with a 50\% probability.
    \item \textbf{Color Jitter:} The brightness and contrast of the images are randomly adjusted by a factor of up to 0.1.
    \item \textbf{Normalization:} Image pixel values are normalized to have a mean of 0.449 and a standard deviation of 0.226.
\end{itemize}

\section{Experiments on LiverCT}
\label{subsec:liver_supp}

We performed a few pre-processing steps on the LiverCT \cite{LiTs, landman2015Altas} benchmark. In this section, we introduce these techniques one by one and complete the Table~\ref{tab:liver_ablation}

\subsection{Data Preprocessing}

\begin{figure}[h]
  \centering
  \includegraphics[width=0.9\linewidth]{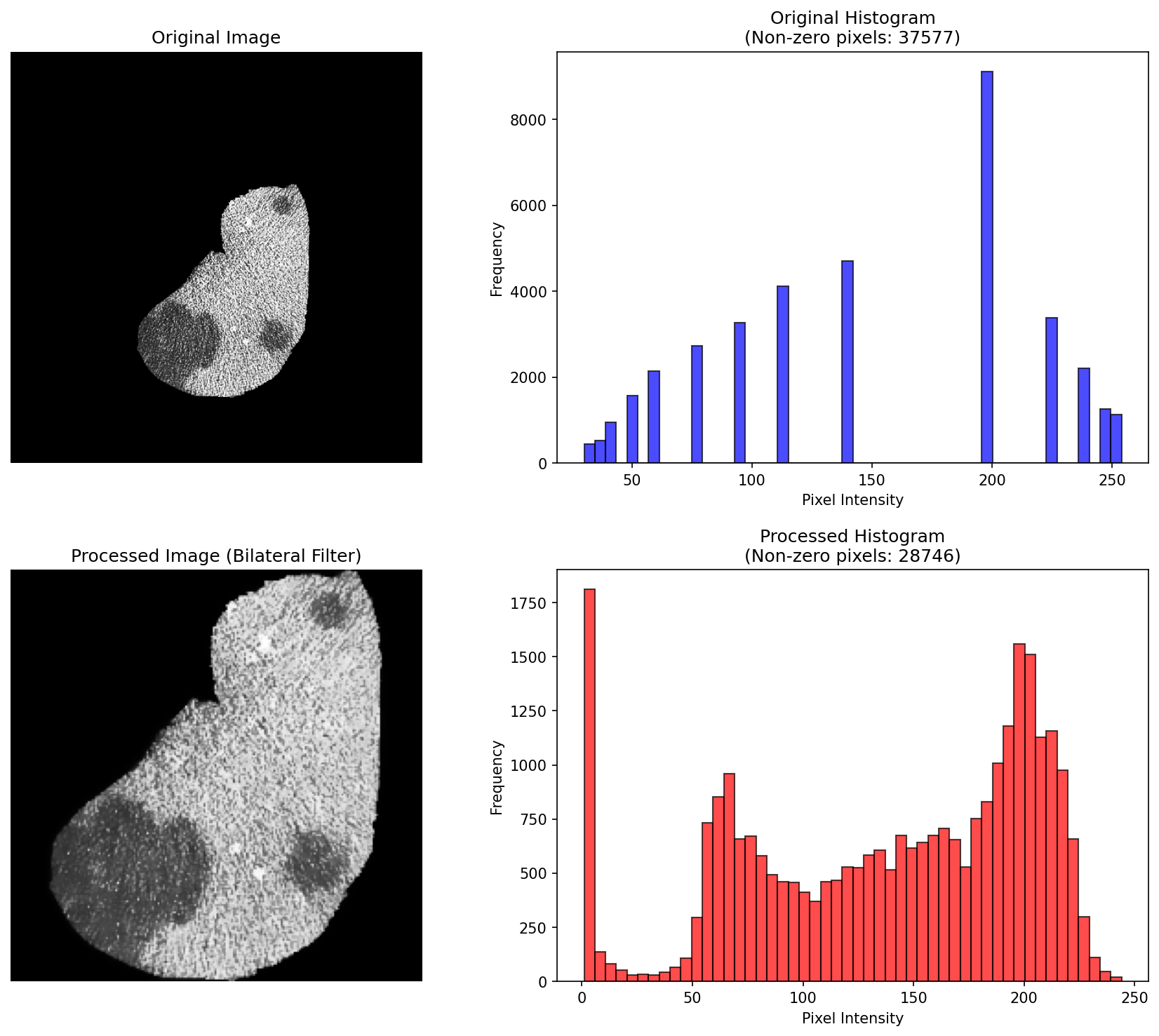}
      
  \caption{Original and processed images from LiverCT \cite{LiTs, landman2015Altas} along with their pixel histograms.}  
  \label{fig:liver_process}
\end{figure}

First of all, the dimensions of images in the dataset are 512x512 pixels, and only small portion of the images have the Liver segments. Resizing these images to 224x224 pixels, which is the input size of our model, results in diminished liver sections. To resize without losing details of the region of interest (i.e. liver section) we used the following algorithm to get images resized to 224x224. Note that this process is fully automated and can be applied to any segmented liver images.

\begin{enumerate}
    \item \textbf{ROI Identification:} For each $512 \times 512$ input image, we first identify the region containing the liver. This is achieved by computing a union bounding box that tightly encloses all non-zero pixels.

    \item \textbf{ROI Cropping:} The image is cropped using the coordinates of the calculated bounding box, isolating the liver segment from the empty background.

    \item \textbf{Canvas Preparation:} A new, black canvas of the target dimensions ($224 \times 224$) is created to serve as the background for the final model input.

    \item \textbf{Conditional Resizing and Placement:} The cropped liver ROI is placed onto the canvas using a size-dependent strategy:
    \begin{itemize}
        \item \textbf{If the ROI is smaller than or equal to $224 \times 224$:} The cropped segment is pasted directly onto the center of the canvas without any resizing. This preserves the native resolution of the liver tissue.
        \item \textbf{If the ROI is larger than $224 \times 224$:} The segment is resized to fit within the $224 \times 224$ frame while maintaining its original aspect ratio to prevent distortion. The resized ROI is then centered on the canvas.
    \end{itemize}

    \item \textbf{Final Input:} The resulting $224 \times 224$ image, with the liver segment prominently centered, is used as the input for the model.
\end{enumerate}

Another issue with this dataset is that, due to constraints inherent to Computed Tomography imaging, it underwent several windowing and histogram equalization techniques \cite{bilic2023liver, bmad, li2022_MICCAI_liver}. As a result, these images can be out of distribution of standard datasets like ImageNet, on which our employed perceptual loss model is trained. To mitigate this, we apply a bilateral filter~\cite{bilateral} to each processed $224 \times 224$ image prior to feeding it to the network.

The effect of the pre-processing is illustrated in Figure \ref{fig:liver_process}, where the ROI and anomalous regions are preserved, and the histogram of the image looks more natural.

Retraining and re-evaluating the model with this new preprocessing algorithm yieded the results in Row \rownumber{2} of the Table \ref{tab:liver_ablation}. This result is mean and standard devation of evaluation of 5 different models trained with 5 different seed (42, 7, 13, 65, 91)

\subsection{New Evaluation Config}

\begin{table}
  \centering
   \caption{Ablations on LiverCT Dataset.}
\resizebox{0.99\linewidth}{!}{
   
  \begin{tabular}{lll}
    \toprule
     & \bf Version &  \bf AUROC  \\
    \midrule
    \rownumber{1} & Main Config \ref{tab:train_hyperparams} & $54.1$  \\
    \rownumber{2} &+ Train $\&$ Eval with New Preprocessing &  $59.5 \pm 1.27$  \\
    \rc \rownumber{3} & \rc + Eval Perceptual Patch Sizes [16, 32, 56] $->$ [8, 16] & \rc $65.5 \pm 1.96$   \\

    \bottomrule
  \end{tabular}
}
 
  \label{tab:liver_ablation}
\end{table}

\begin{figure}[h]
  \centering
  \includegraphics[width=0.99\linewidth]{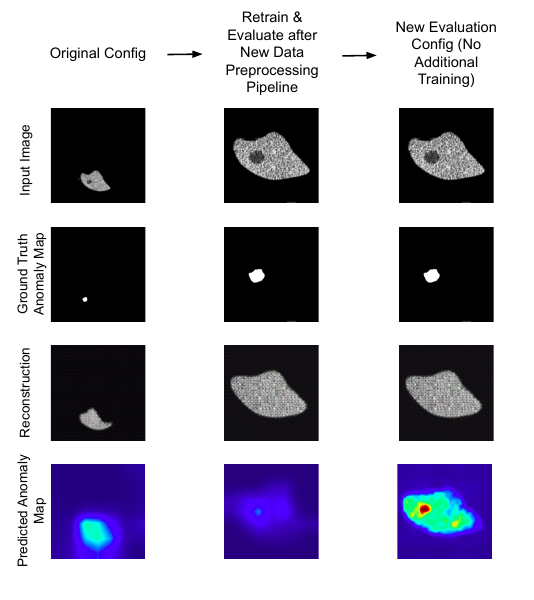}
      
  \caption{Effect of the modifications such as data preprocessing pipeline and evaluation config. We first avoid diminishing the anomalous region during resizing. Then configured perceptual loss to be more biased towards textual clues following insights from literature on visual perception.}
  \label{fig:liver_ablation}
\end{figure}

As shown in Figure \ref{fig:liver_ablation}, the new data preprocessing pipeline (column 2) has improved over the original config (column 1) regarding the quality of the anomaly map and made the anomalous region more visible. However, the predicted anomaly map still fails to capture the texture change in the anomalous region. Following the studies on visual perception \cite{naseer2021intriguingpropertiesvisiontransformers, raghu2022visiontransformerslikeconvolutional} that state smaller patch sizes are biased towards textures while larger patch sizes are biased towards shape, we changed the patch sizes used by perceptual model for anomaly score calculation from [16, 32, 56] to [8, 16]. As can be seen from the third column of Table~\ref{fig:liver_ablation}, with this evaluation config anomaly maps capture texture changes on anomalous regions better. This reflects on the AUROC score of the \rownumber{3}rd row of the Table \ref{tab:liver_ablation}. The evaluation configuration that yields the best result on LiverCT is presented in Table~\ref{tab:eval_hyperparams_best_liver}, with the modified parts highlighted in bold. The training config is kept the same.

\begin{table}[h!]
\centering
\caption{Best Evaluation Configuration on LiverCT.}
\label{tab:eval_hyperparams_best_liver}
\resizebox{\columnwidth}{!}{%
\begin{tabular}{@{}lll@{}}
\toprule
\textbf{Component} & \textbf{Parameter} & \textbf{Value} \\ \midrule
\textbf{General} & Batch Size & 64 \\
 & Test Data Augmentation & None (only resize and normalize) \\ \midrule
\multirow{4}{*}{\textbf{Perceptual Metric}} & Pre-trained Perceptual Model & MAE with ViT-Large Encoder \\
 & Distance Metric & Cosine Distance \\
 & Layers Used for Feature Extraction & From the 12th, 16th, and 20th transformer blocks \\
 & Multi-Scale Input Patch Sizes & \textbf{8x8x, 16x16} \\ \midrule
\multirow{2}{*}{\textbf{Image-Level Score Aggregation}} & Spatial Aggregation per Feature Map & Max \\
 & Cross-Feature Map Aggregation & Mean \\ \midrule
\multirow{1}{*}{\textbf{Pixel-Level Map Aggregation}} & Cross-Feature Map Aggregation & Mean \\ \bottomrule
\end{tabular}%
}
\end{table}

\section{Different Aggregation for Chest RSNA}
\label{subsec:chest_supp}

\begin{figure}[h]
  \centering
  \includegraphics[width=0.3\linewidth]{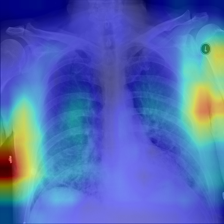}%
  \hfill
  \includegraphics[width=0.3\linewidth]{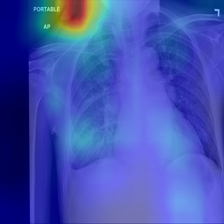}%
  \hfill
  \includegraphics[width=0.3\linewidth]{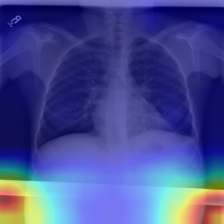}
      
  \caption{Optical characters and artifacts dominate the response from the anomalous region.}
  \label{fig:chest_artifacts}
\end{figure}

It is usual to see different optical characters and artifacts on Chest images. Their position varies. When these artifacts are present, they dominate the anomaly signals, and the anomaly score cannot be calculated properly with the aggregation method described in Main Eq. 4. Since their positions vary and are unpredictable, we were unable to devise a preprocessing algorithm.

To mitigate this problem, we decided to experiment with different aggregation methods on the validation split of the Chest RSNA dataset. As an alternative, we first tried taking the mean value in the anomaly map from each location, and then taking the maximum across different layers. We observed an increase in AUROC from 78.6\% to 84.3\% on the validation split. Therefore, we decided to keep this approach and reported an AUROC of 83.8\% on test set as in main Table 5. The evaluation configuration that yields the best result on Chest RSNA is presented in \ref{tab:eval_hyperparams_best_chest}, with the modified parts highlighted in bold. The training config is kept the same.

\begin{table}[h!]
\centering
\caption{Best Evaluation Configuration on Chest RSNA}
\label{tab:eval_hyperparams_best_chest}
\resizebox{\columnwidth}{!}{%
\begin{tabular}{@{}lll@{}}
\toprule
\textbf{Component} & \textbf{Parameter} & \textbf{Value} \\ \midrule
\textbf{General} & Batch Size & 64 \\
 & Test Data Augmentation & None (only resize and normalize) \\ \midrule
\multirow{4}{*}{\textbf{Perceptual Metric}} & Pre-trained Perceptual Model & MAE with ViT-Large Encoder \\
 & Distance Metric & Cosine Distance \\
 & Layers Used for Feature Extraction & From the 12th, 16th, and 20th transformer blocks \\
 & Multi-Scale Input Patch Sizes & 16x16, 32x32, 56x56 \\ \midrule
\multirow{2}{*}{\textbf{Image-Level Score Aggregation}} & Spatial Aggregation per Feature Map & \textbf{Mean} \\
 & Cross-Feature Map Aggregation & \textbf{Max} \\ \midrule
\multirow{1}{*}{\textbf{Pixel-Level Map Aggregation}} & Cross-Feature Map Aggregation & Mean \\ \bottomrule
\end{tabular}%
}
\end{table}

\section{SOTA Results on Each Dataset}

\begin{table}[h!]
\centering
\caption{Best Training Configuration for Brain MRI.}
\label{tab:train_hyperparams_best_brain}
\resizebox{\columnwidth}{!}{%
\begin{tabular}{@{}lll@{}}
\toprule
\textbf{Component} & \textbf{Parameter} & \textbf{Value} \\ \midrule
\textbf{General} & Seed & 42, 7, 13, 65, 91 (mean of 5 runs are reported)\\
 & Image Resolution (Resize) & 224x224 \\
 & Batch Size & 64 \\
 & Epochs & 300 \\
 & Device & CUDA \\ \midrule
\textbf{Encoder} & Pre-trained Model & \textbf{ViT-L/14 + ViT-B/8} \\
 & Pre-training Method & \textbf{DINOv2 + DINO} \\
 & Frozen During Training & \textbf{True, True} \\
 & Hidden States Used & Features from the 2nd and 4th to last blocks \\
 & Final Projection In-Features & \textbf{1024, 768} \\
 & Final Projection Out-Features & \textbf{768, 768} \\ \midrule
\textbf{Q-Former (Junction)} & Number of Transformer Blocks & 1 \\
 & Internal Dimension & 768 \\
 & Output Dimension & 768 \\
 & Number of Learnable Queries & 784 (for 28x28 output patches) \\
 & Attention Heads & 8 \\
 & MLP Expansion Ratio & 4.0 \\ \midrule
\textbf{Decoder} & Internal Dimension & 768 \\
 & Depth (Number of Layers) & 6 \\
 & Attention Heads & 12 \\
 & Output Patch Size & 8x8 \\
 & Number of Output Patches & 28x28 \\
 & MLP Expansion Ratio & 4.0 \\ \midrule
\textbf{Optimization} & Optimizer & Adam \\
 & Learning Rate (Maximum) &  $8 \times 10^{-5}$ \\
 & Learning Rate Scheduler & OneCycleLR \\ \midrule
\multirow{4}{*}{\textbf{Perceptual Loss}} & Pre-trained Perceptual Model & MAE with ViT-Large Encoder \\
 & Distance Metric & Cosine Distance \\
 & Layers Used for Feature Extraction & From the 16th and 20th transformer blocks \\
 & Multi-Scale Input Patch Sizes & 32x32, 56x56 \\ \bottomrule
\end{tabular}%
}
\end{table}



As shown in Table \ref{tab:final_results}, we achieve the state-of-the-art performance in BraTS2021 ~\cite{baid2021rsna, bakas2017advancing, menze2014multimodal}, RESC~\cite{hu2019automated} and RSNA~\cite{wang2017chestx} and second on LiverCT~\cite{LiTs, landman2015Altas}.

\begin{table}[t]
    \centering
    \renewcommand{\arraystretch}{1.5}  
       
    \caption{Anomaly detection performance (mean + std) on BraTS2021, Liver CT (BTCV + LiTs), RESC and RSNA. The results are reported for five repetitions of the experiment. $^*$:~denotes only three repetitions. The top results are reported in bold.
    }
    \resizebox{0.99\linewidth}{!}{
    \begin{tabular}{l|cccc}
        \toprule
        \bf  Methods & \bf BraTS2021 & \bf Liver CT & \bf RESC &  \bf RSNA \\ 
        \midrule
         f-AnoGAN~\citep{schlegl2019f} & $77.3 \pm 0.18$  & $58.4 \pm 0.15$ & $77.4 \pm 0.85$  &  $55.6 \pm 0.09$  \\
         GANomaly~\citep{akcay2018ganomaly}& $74.8 \pm 1.93$ & $53.9 \pm 2.36$ & $52.6 \pm 3.95$ & $62.9\pm 0.65$  \\
         DRAEM~\citep{zavrtanik2021draem} & $62.4 \pm 9.03$ & $\mathbf{69.2 \pm 3.86}$ & $83.2 \pm 8.21$ & $67.7 \pm 1.72$   \\
             UTRAD~\citep{chen2022utrad}  & $82.9 \pm 2.32$  & $55.6 \pm 5.96$ & $89.4 \pm 1.92$ & $75.6 \pm 1.24$ \\ 
         
        DeepSVDD~\citep{ruff2018deep}  & $87.0 \pm 0.66$ & $53.3 \pm 1.24$ & $74.2 \pm 1.29$  & $64.5 \pm 3.17$ \\
             CutPaste~\citep{li2021cutpaste}  & $78.8 \pm 0.67$  & $58.6 \pm 4.2$ & $90.2 \pm 0.61$  & $82.6 \pm 1.22$ \\
             SimpleNet~\citep{liu2023simplenet} & $82.5 \pm 3.34$ & N/A & $76.2 \pm7.46$ & $69.1 \pm 1.27$ \\
             
             MKD~\citep{salehi2021multiresolution} & $81.5 \pm 0.36$  & $60.4 \pm 1.61$ &  $89.0 \pm 0.25$  &  $82.0 \pm 0.12$ \\
             
             RD4AD~\citep{deng2022anomaly} &  $89.5 \pm 0.91$ & $60.0 \pm 1.4$ &$87.8 \pm 0.87$ &  $67.6 \pm 1.11$ \\
            
             STFPM~\citep{yamada2021reconstruction}  & $83.0 \pm 0.67$ & $61.6 \pm 1.7$ &  $84.8 \pm 0.50$  & $72.9 \pm 1.96$ \\
             PaDiM ~\citep{defard2021padim} & $ 79.0 \pm 0.38 $& $50.7 \pm 0.5$ & $75.9 \pm0.54$  & $77.5 \pm 1.87$  \\
             PatchCore~\citep{roth2022towards}  & $91.7 \pm 0.36$  & $60.4 \pm 0.82$ &  $91.6 \pm0.10$  & $76.1 \pm 0.67$\\
             CFA~\citep{lee2022cfa}  & $84.4 \pm 0.87$ & $61.9 \pm 1.16$ & $69.9 \pm 0.26$ &  $66.8 \pm 0.23$ \\
            CFLOW~\citep{gudovskiy2022cflow} & $74.8 \pm 5.32$  & $49.9 \pm 4.67$ & $75.0 \pm 5.81$  & $71.5 \pm 1.49$    \\
             CS-Flow~\citep{rudolph2022fully}  & $90.9 \pm 0.83$ & $59.4 \pm 0.52$ &  $87.3 \pm 0.58$ & $83.2 \pm 0.46$  \\ 
             P-VQ$^*$~\citep{pvq}  &  $94.3 \pm 0.23$ & $60.6 \pm 0.62$ &  $89.0 \pm 0.48$ & $79.2 \pm 0.04$  \\ 
              \rc \bf QFAE (ours)  & \rc $\mathbf{94.3 \pm 0.18}$ & \rc $65.5 \pm 1.96$ & \rc   $\mathbf{91.8 \pm 0.55}$ & \rc   $\mathbf{83.8 \pm 0.46}$  \\ 
        \bottomrule
    \end{tabular}
  }
    \label{tab:final_results}
\end{table}